\documentclass[10pt,twocolumn,letterpaper]{article}

\usepackage{iccv}              

\usepackage{multirow}
\usepackage{afterpage}

\definecolor{iccvblue}{rgb}{0.21,0.49,0.74}
\usepackage[pagebackref,breaklinks,colorlinks,allcolors=iccvblue]{hyperref}

\def\paperID{10055} 
\def\confName{ICCV}
\def\confYear{2025}

\title{Seam360GS: Seamless 360° Gaussian Splatting \\ from Real-World Omnidirectional Images}
\author{Changha Shin \hspace{12pt} Woong Oh Cho \hspace{12pt} Seon Joo Kim\\
Yonsei University\\
{\tt\small \{changhashin, wocho, seonjookim\}@yonsei.ac.kr}
}

\begin{document}
\twocolumn[{
\maketitle
\vspace{-1.4cm}
\begin{center}
    \includegraphics[width=0.90\linewidth]{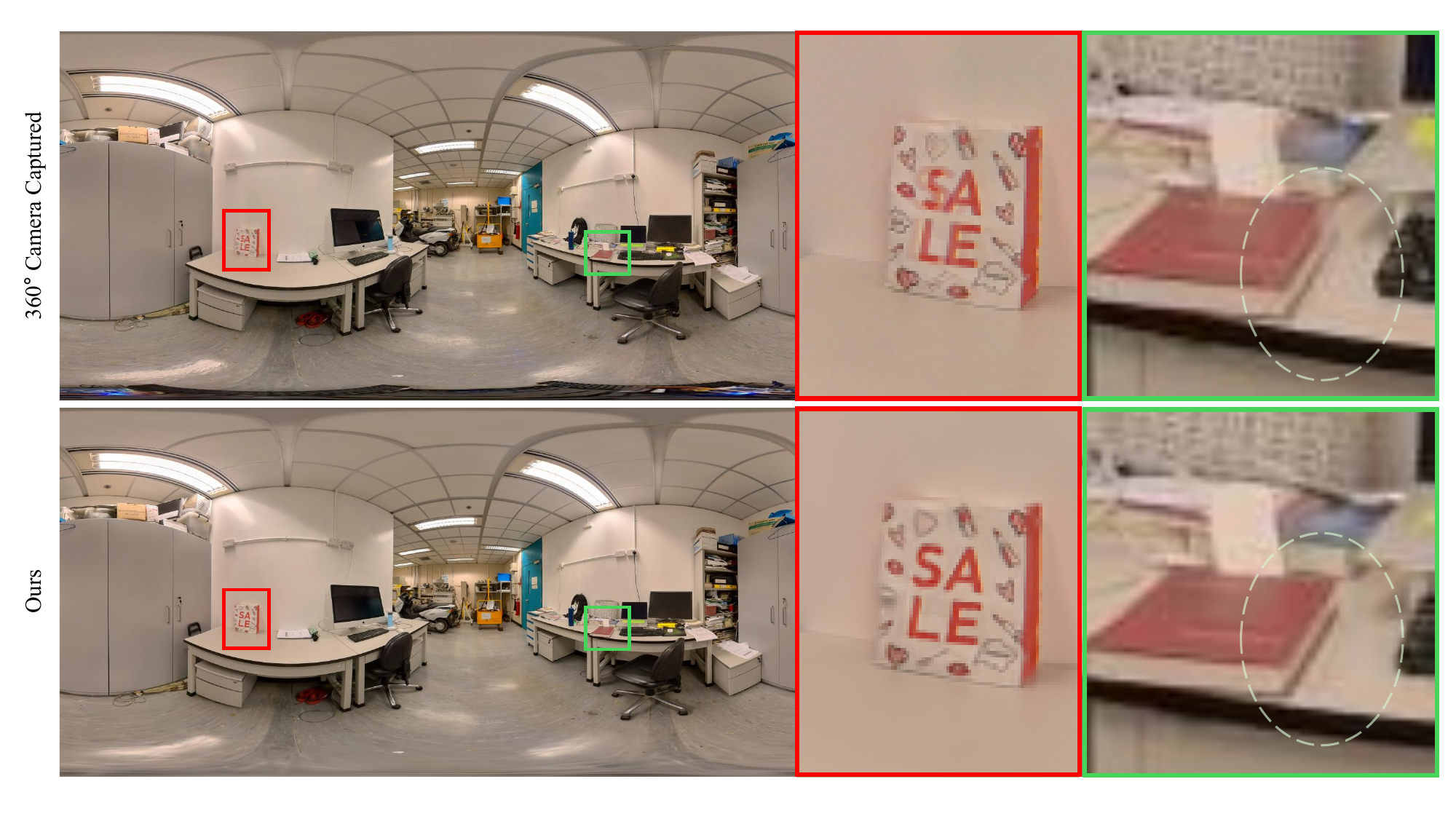}
    \vspace{-0.5cm}
    \captionof{figure}{(Top) An indoor scene captured by a 360° camera, with red and green boxes indicating regions affected by stitching artifacts. (Bottom) Our model's novel view synthesis result, demonstrating seamless rendering of scene details—including text and objects—free from stitching artifacts, with overall improvements in rendering quality and consistency.
    }
    \label{Fig:figure0_headline}
\end{center}
}]

\begin{abstract}
360° visual content is widely shared on platforms such as YouTube and plays a central role in virtual reality, robotics, and autonomous navigation. However, consumer-grade dual-fisheye systems consistently yield imperfect panoramas due to inherent lens separation and angular distortions. In this work, we introduce a novel calibration framework that incorporates a dual-fisheye camera model into the 3D Gaussian splatting pipeline. Our approach not only simulates the realistic visual artifacts produced by dual-fisheye cameras but also enables the synthesis of seamlessly rendered 360° images. By jointly optimizing 3D Gaussian parameters alongside calibration variables that emulate lens gaps and angular distortions, our framework transforms imperfect omnidirectional inputs into flawless novel view synthesis. Extensive evaluations on real-world datasets confirm that our method produces seamless renderings—even from imperfect images—and outperforms existing 360° rendering models.
\end{abstract}    
\vspace{-0.2cm}
\section{Introduction}
\label{sec:intro}

\hspace{0.4cm}Omnidirectional cameras have become integral to applications such as virtual reality, robotics, and autonomous navigation due to their ability to capture the entire surrounding environment in a single shot. These devices offer an expansive field of view and rich spatial context, both of which are crucial for robust scene understanding and high-level processing. However, most consumer-grade systems (e.g., Insta360, Ricoh360 \cite{insta360, ricoh360}) rely on dual fisheye lens configurations, which inherently introduce challenges. The physical separation between lenses and the angular distortions from the image stitching process result in panoramas that deviate from ideal geometric calibration, as illustrated in the top panel of \cref{Fig:figure0_headline}. These calibration inaccuracies can impair downstream tasks that depend on precise spatial reconstructions.

Recent advances in 3D reconstruction exemplified by Neural Radiance Fields (NeRF)~\cite{mildenhall2021nerf} and 3D Gaussian splatting (3DGS)~\cite{kerbl20233dgs} have enabled photorealistic novel view synthesis. NeRF employs deep networks to learn a continuous radiance field from multi-view images, while 3DGS represents a scene using a sparse set of anisotropic Gaussians, allowing for efficient, real-time rendering. Since these methods were primarily designed for perspective images, they often fail to produce accurate renderings for omnidirectional data.
To address these challenges, several works~\cite{bai2024360gs, lee2024odgs, li2024omnigs, wang2024op43dgs} extend 3DGS to reconstruct an omnidirectional radiance field by projecting 3D Gaussians onto the tangent planes of a unit sphere. However, these approaches generally presume that the panoramic images have been accurately calibrated through spherical mapping.

Numerous studies~\cite{nerfmm2021, scnerf2021, lin2021barf, park2023camp, sparf2022sparf, doe2022colmapfree} have sought to improve novel view synthesis through camera calibration; however, most of these methods struggle with omnidirectional images having extremely wide fields of view. A recent 3DGS-based approach~\cite{huang2025scomnigs} calibrates lens distortions in 360° images while refining camera poses. However, it addresses only camera distortions and neglects the inter-camera gap in dual fisheye systems—an omission that significantly constrains seamless image synthesis.

In this paper, we propose an integrated calibration framework that jointly optimizes both the 3D Gaussian parameters and additional calibration variables to compensate for lens gap separation and angular distortions. Specifically, we redefine Gaussian splat positions by translating them to correspond with the optical centers of the fisheye cameras and apply 3D transformations to simulate rotational distortions introduced by the camera rays. This strategy refines the reconstructed omnidirectional radiance field and facilitates seamless 360° image rendering, as shown in the bottom panel of \cref{Fig:figure0_headline}, all while maintaining the efficiency of the Gaussian Splatting paradigm. In summary, our contributions are:
\begin{itemize}[leftmargin=*]

    \vspace{0.3em}\item We propose a novel framework that integrates omnidirectional dual-fisheye distortion modeling with the 3D Gaussian splatting paradigm. Our algorithm optimizes the ideal 3D positions of Gaussian splats by jointly calibrating for both lens distortions and the inter-camera gap.
    
    \vspace{0.3em}\item Our approach enables seamless omnidirectional novel view synthesis without incurring additional computational overhead. During training, we generate realistic, imperfect 360° images that simulate dual fisheye artifacts and compare them to actual captures to learn robust calibration parameters. At inference, the calibration module is deactivated, ensuring that the system operates efficiently while still delivering flawlessly rendered 360° views.
     
    
    \vspace{0.3em}\item Extensive experiments on real-world 360° benchmark datasets \cite{choi2023egonerf, huang2022360roam} demonstrate that our method consistently outperforms existing state-of-the-art algorithms for omnidirectional novel view synthesis.
\end{itemize}

\vspace{-0.1cm}
\section{Related Works}
\label{sec:relatedworks}

\subsection{Classical Image Stitching Approaches}
Classical omnidirectional image stitching techniques produce seamless panoramic images from overlapping views captured by fisheye or dual-fisheye cameras. Early methods followed a sequential pipeline starting with rigorous calibration to correct the pronounced radial distortions inherent in wide-angle lenses. By accurately modeling these distortions, the input images were normalized to establish a robust foundation for subsequent feature extraction and alignment \cite{Brown07,Lowe04}. In this pipeline, robust keypoints were detected using descriptors such as SIFT \cite{Lowe04} or SURF \cite{Bay08} and then matched across overlapping regions, with RANSAC \cite{Fischler81} used to compute global homographies under the assumption of planar scenes—a premise that often fails in environments with significant parallax and depth variations.

To address these challenges, later research introduced multi-homography models and spatially varying warping strategies that adapt transformations locally to better accommodate non-planar geometries. Advanced blending techniques were concurrently developed to smooth intensity variations and eliminate visible seams, ensuring consistent brightness and color continuity. However, these traditional stitching pipelines generally assume near-ideal capture conditions and sufficient overlap, which can result in artifacts when applied to consumer-grade dual-fisheye systems.

More recently, deep learning approaches have emerged to overcome these limitations by learning robust, high-level representations that directly tackle issues such as parallax, non-planarity, and exposure mismatches in dual-fisheye and 360° imaging systems. For instance, the OmniStitch \cite{omnistitch2024} integrates depth cues via a depth‐aware network, enabling more reliable alignment across varying distances and complex geometries while reducing ghosting and misregistration artifacts. Similarly, attentive deep stitching methods \cite{attentive2020} employ attention mechanisms to adaptively fuse overlapping regions, effectively managing local intensity differences and subtle misalignments, while additional depth-aware strategies \cite{depthaware2023} further enhance robustness under large parallax conditions. Comprehensive surveys on deep learning for omnidirectional vision \cite{Ai2022} highlight that these advances relax many of the restrictive assumptions of classical pipelines, achieving superior visual consistency and artifact suppression. Nonetheless, despite producing perceptually appealing panoramas, deep learning–based methods remain constrained by the inherent physical limitations of the imaging process, often resulting in suboptimal geometric fidelity for applications demanding rigorous spatial accuracy.

\subsection{Novel View Synthesis for 360° Imaging}
Neural Radiance Fields (NeRF) \cite{mildenhall2021nerf} have revolutionized novel view synthesis by representing scenes as continuous volumetric radiance fields parameterized by neural networks. However, NeRF typically requires extensive training time and significant computational resources, prompting research into more efficient representations. In this context, 3D Gaussian splatting has emerged as a compelling alternative. By modeling a scene as a collection of anisotropic Gaussian primitives, this approach offers an explicit, point-based representation that can be rendered in real time while maintaining competitive visual quality. Unlike the implicit volumetric representations in NeRF, 3D Gaussian splatting leverages well-localized Gaussian functions to approximate scene geometry and appearance, thereby enabling rapid synthesis of novel views with significantly reduced rendering costs.

\vspace{0.1cm}
\noindent{\bf NeRF-Based Approaches.} 
While the foundational NeRF framework has already been described in earlier works, its extension to omnidirectional data has spawned a number of specialized approaches tailored for 360° imagery. Methods such as \cite{huang2022360roam, kulkarni2023fusion, choi2023egonerf, chen2023panogrff} adapt the standard NeRF pipeline to accommodate the challenges posed by panoramic inputs. These approaches modify the conventional sampling and representation strategies to work in spherical coordinates, thereby effectively handling the wide field-of-view inherent to 360° cameras.
For example, 360Roam~\cite{huang2022360roam} employs geometry-aware sampling to capture indoor environments robustly, while 360FusionNeRF~\cite{kulkarni2023fusion} and PanoGRF~\cite{chen2023panogrff} extend the NeRF paradigm to support wide-baseline panoramas by integrating specialized feature grids and joint guidance strategies. Notably, EgoNeRF~\cite{choi2023egonerf} focuses on egocentric scenes by utilizing a quasi-uniform spherical grid that adapts to the unbounded nature of panoramic views. Notably, while these approaches generally deliver impressive reconstruction quality, 360Roam~\cite{huang2022360roam} points out a potential limitation: manufacturing imperfections and suboptimal factory calibration can introduce noticeable stitching artifacts in the source images, which may adversely affect the final rendering quality.

\vspace{0.1cm}
\noindent{\bf 3D Gaussian Splatting Approaches.} 
Recent advances in omnidirectional radiance field reconstruction—exemplified by works such as 360GS \cite{bai2024360gs}, ODGS \cite{lee2024odgs}, OP43DGS \cite{wang2024op43dgs}, and OmniGS \cite{li2024omnigs} extend the 3D Gaussian Splatting framework to 360° imagery by leveraging specialized rasterizers that project 3D Gaussians onto tangent planes of a unit sphere.
360GS adapts the standard splatting technique for optimal panoramic rendering, ODGS enhances fine detail recovery through directional augmentation, OP43DGS employs an optimized splatting approach to accurately reconstruct depth and complex scene structures, and OmniGS leverages the optical properties of omnidirectional images to achieve fast, precise radiance field reconstruction.

While most NeRF and 3DGS methods assume that mapping omnidirectional images to a spherical domain ensures inherently accurate internal geometry, our work challenges this notion by showing that dual fisheye camera systems introduce intrinsic calibration issues. Specifically, a gap between the two fisheye lenses and the resulting angular distortions in the camera rays deviate from the idealized geometric model.

\subsection{Neural Rendering with Camera Calibration}
Recent research in neural rendering has increasingly focused on jointly optimizing scene representations and camera parameters to overcome the limitations imposed by imperfect calibration. For instance, NeRF\symbol{45}\symbol{45} \cite{nerfmm2021} was one of the earliest works to jointly optimize scene representations and camera parameters, thereby eliminating the need for pre-calibrated inputs. Building on this foundation, methods such as SCNeRF \cite{scnerf2021} integrate both intrinsic and extrinsic calibration into the NeRF optimization process by simultaneously refining camera intrinsics (e.g., focal length and distortion coefficients) and extrinsics (e.g., camera poses) to compensate for lens imperfections and misalignments common in consumer-grade imaging systems. Other approaches, including BARF \cite{lin2021barf}, leverage bundle adjustment to iteratively refine camera poses, while additional works like Nope-NeRF \cite{nope2022nope}, CAMP \cite{park2023camp}, and SPARF \cite{sparf2022sparf} incorporate advanced pose estimation and sparse representation techniques for efficient view synthesis. Collectively, these methods demonstrate a powerful trend toward reducing reliance on pre-calibrated inputs by addressing calibration errors during training, thereby enhancing the overall quality of novel view synthesis.

Similar calibration strategies have been extended to alternative neural rendering approaches such as 3D Gaussian Splatting. In particular, COLMAP-Free 3D Gaussian Splatting \cite{doe2022colmapfree} has emerged that eliminates the dependency on external Structure-from-Motion pipelines. These approaches embed calibration corrections within the Gaussian splatting framework itself, jointly optimizing both the scene representation and the camera parameters. This integration not only streamlines the processing pipeline but also enhances robustness against residual calibration errors, whether they originate from intrinsic distortions or extrinsic misalignments. Collectively, these calibrated neural rendering methods demonstrate a powerful trend towards reducing reliance on pre-calibrated inputs. By addressing both intrinsic and extrinsic errors during training, these methods achieve improved alignment and rendering quality, a critical advancement for challenging scenarios such as 360° novel view synthesis from dual-fisheye cameras.

Recent works have also focused on calibrating 360° images for neural rendering. For example, Omni-NeRF \cite{gu2022omninerf} leverages raw dual fisheye images to reconstruct neural radiance fields, primarily mitigating wide field-of-view lens distortions, yet it does not address the stitching artifacts in omnidirectional images commonly observed in consumer-grade systems. Additionally, most methods rely on equirectangular images that have been pre-corrected, which makes obtaining raw fisheye images challenging in many cases. 
SC-OmniGS \cite{huang2025scomnigs} jointly optimizes both lens distortions and camera poses to enhance the quality of the reconstructed scene. It effectively addresses lens distortions using a UV mapping approach, correcting them directly within the 2D image domain. However, this method primarily focuses on lens distortions and does not account for inter-camera gaps. In contrast, our algorithm explicitly adjusts the positions of Gaussian splats in the 3D scene space by jointly calibrating for both lens distortions and inter-camera gaps, leading to more accurate geometric reconstruction and seamless omnidirectional rendering.

\section{Methodology}

In this section, we present the preliminaries on 3D Gaussian splatting and its extension to 360° imaging, and then describe our integrated calibration framework.
\subsection{Preliminaries: 3D Gaussian splatting}

3D Gaussian splatting represents a scene as a set of anisotropic Gaussian primitives, each of which encodes local geometric and appearance information. A single Gaussian primitive is defined by its center, covariance, and associated color and opacity. Formally, a Gaussian is given by:
\begin{equation}
    G(\mathbf{x}) = \exp\left(-\frac{1}{2} (\mathbf{x} - \boldsymbol{\mu})^T \Sigma^{-1} (\mathbf{x} - \boldsymbol{\mu})\right),
    \label{eq:gaussian}
\end{equation}
where \(\boldsymbol{\mu} \in \mathbb{R}^3\) is the center of the Gaussian and \(\Sigma \in \mathbb{R}^{3 \times 3}\) is the covariance matrix that captures its anisotropic spread.

During rendering, the 3D Gaussians are projected onto the image plane via a camera projection function \(\pi(\cdot)\). The projected Gaussians are then sampled and \(\alpha\)-blended to produce the final image. In our simplified formulation, the image intensity at pixel \(p\) is computed as
\begin{equation}
    I(p) = \sum_{i} \alpha_i(p)\, c_i,
\end{equation}
where the blending weight for the \(i\)th Gaussian is defined as
\begin{equation}
    \alpha_i(p) = o_i\, \exp\!\Bigl(-\frac{1}{2}\|p - \pi(m_i)\|_{W_i}^2\Bigr).
\end{equation}
Here, \(o_i\) is the opacity of the \(i\)th Gaussian, \(\pi(m_i)\) denotes its projected center, and \(\|p - \pi(m_i)\|_{W_i}^2\) represents the Mahalanobis distance with respect to the projected covariance \(W_i\). This differentiable forward rendering pipeline enables the joint optimization of all Gaussian parameters through photometric loss, even under sparse sampling conditions.

To adapt 3D Gaussian splatting for omnidirectional imaging, \cite{li2024omnigs, lee2024odgs, bai2024360gs, wang2024op43dgs} project the Gaussian primitives onto a unit sphere, normalizing each 3D point’s coordinates to capture a full 360° view of the scene. 
While extending 3D Gaussian splatting to omnidirectional imaging establishes an effective rasterizer by projecting Gaussian primitives onto a unit sphere, their approach inherently assumes that the input 360° images are accurately stitched. These imperfections degrade geometric accuracy and visual consistency in the rendered omnidirectional outputs.


\subsection{Dual-Fisheye Distortion Modeling and Calibration for 3D Gaussian splatting}

\begin{figure*}
    \centering
    \includegraphics[width=0.93\linewidth, trim=0 122pt 0 30pt, clip]{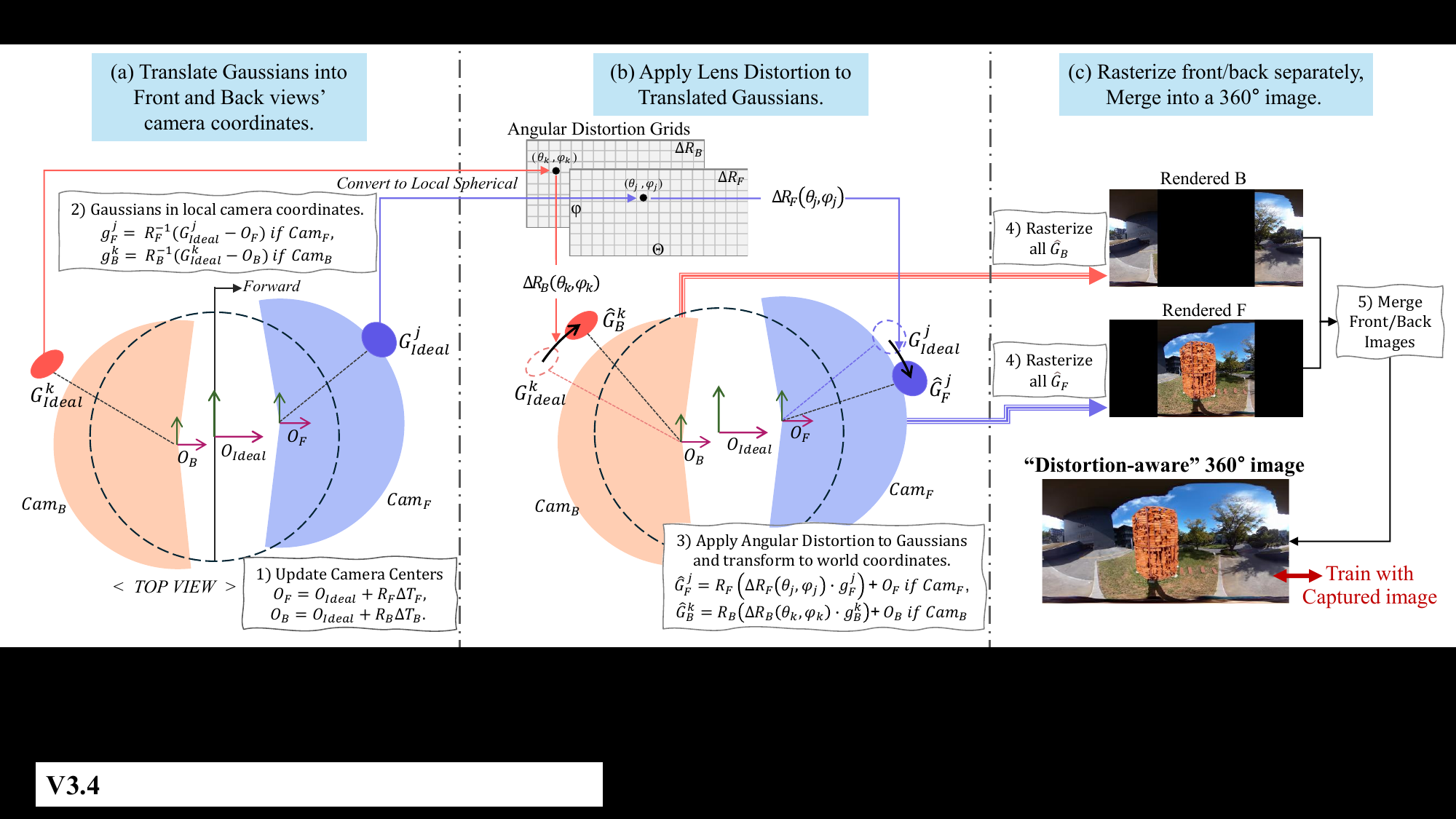}
    \vspace{-0.2cm}
    \caption{Overview of our seamless 360° Gaussian Splatting rendering pipeline, which integrates fisheye camera stitching and applies lens distortion to fully replicate the distortions observed in actual camera captures.}
    \vspace{-0.3cm}
    \label{Fig:GS_figure_01}
\end{figure*}

To simulate the inherent imperfections of dual-fisheye systems, we convert the ideal Gaussian primitives into camera-specific representations via two key steps: (1) translating the Gaussians to align with the dual-fisheye centers, and (2) applying angular distortion to emulate errors arising in the omnidirectional stitching process. A schematic overview of our proposed algorithm is illustrated in \cref{Fig:GS_figure_01}.

\vspace{0.1cm}
\noindent{\bf Translation of Gaussians to Dual-Fisheye Centers.} 
In practical dual-fisheye systems, hardware constraints lead to a misalignment between the optical centers of the front and back cameras, preventing them from coinciding with an idealized camera center. To correct for this discrepancy, we incorporate a calibration step that translates the Gaussian primitives accordingly as shown in Fig.~\ref{Fig:GS_figure_01}(a). Specifically, we denote the front and back camera centers as \(O_{\text{F}}\), \(O_{\text{B}}\) respectively, relative to an ideal center \(O_{\text{ideal}}\). This update is performed using learnable translation vectors \(\Delta T_F\) and \(\Delta T_B\), which are optimized via backpropagation during the Gaussian Splatting training process.
Specifically, the updated camera centers are given by
\begin{equation}
O_F = O_{\text{ideal}} + R_F\,\Delta T_F,
\end{equation}
\begin{equation}
O_B = O_{\text{ideal}} + R_B\,\Delta T_B,
\end{equation}
where \(R_F\) and \(R_B\) are the rotation matrices corresponding to the front and back cameras.

Each Gaussian, originally defined in the ideal camera coordinate system as \(G_{\text{ideal}}\), is translated into the dual-fisheye camera coordinates. For the front and back cameras, we have
\begin{equation}
g_F^j = R_F^{-1}\bigl(G_{\text{ideal}} - O_F\bigr),
\end{equation}
\begin{equation}
g_B^k = R_B^{-1}\bigl(G_{\text{ideal}} - O_B\bigr).
\end{equation}
Here, the index \(j\) denotes each Gaussian element associated with the front camera, while \(k\) denotes each Gaussian element associated with the back camera.

\noindent{\bf Simulated Angular Distortion in Omnidirectional Image Stitching.} 
\hspace{0.2cm}To account for the angular distortions introduced during the stitching process in dual-fisheye systems, we apply angular distortions to the Gaussian primitives. In our formulation, the angular distortion parameter \(\Delta\)$R$ is modeled as a learnable embedding. For the front camera, this embedding is denoted as \(\Delta\)$R_F \in \mathbb{R}^{M \times N \times 3}$, where \(M\) and \(N\) represent the discretized resolutions along the spherical coordinates \(\theta\) and \(\phi\), respectively, and the final dimension (of size 3) encodes the corresponding rotation adjustments. As illustrated in \cref{Fig:GS_figure_01}(b), given a Gaussian with spherical coordinates \((\theta, \phi)\) in its local camera coordinate system, bilinear sampling from \(\Delta\)$R_F$ yields a rotation offset that specifies the degree to which the Gaussian should be rotated—relative to the local camera frame—to model the stitching-induced angular distortions, satisfying \(\Delta R_F(\theta_j, \phi_j) \in \mathrm{SO}(3)\), which is applied in Eqs.~(8)--(9) to model ray-dependent distortion. Similarly, a learnable embedding \(\Delta\)$R_B$ is defined for the back camera. After translating each Gaussian into its local camera coordinate system, these angular distortion adjustments are applied, resulting in distorted Gaussian representations that more accurately reflect the true scene geometry.


To align the translated Gaussians with the spherical representation of each camera’s field, we apply angular distortions \(\Delta R_F(\theta,\phi)\) and \(\Delta R_B(\theta,\phi)\). The corrected Gaussians are given by
\begin{equation}
G_F^j = R_F\left(\Delta R_F(\theta_j,\phi_j) \cdot g_F^j\right) + O_F,
\end{equation}
\begin{equation}
G_B^k = R_B\Bigl(\Delta R_B(\theta_k,\phi_k) \cdot g_B^k\Bigr) + O_B.
\end{equation}
These distortions, inherent to the dual-fisheye setup, integrate the omnidirectional image stitching process directly into the 3D Gaussian splatting framework, enabling the realistic and seamless rendering of 360° images despite calibration imperfections.

The Transformed Gaussians are rasterized within each camera’s view using a 360° Gaussian rasterizer as depicted in \cref{Fig:GS_figure_01} (c). For the front and back cameras, the Gaussian primitives are independently projected to generate equirectangular images, which are then concatenated to form a complete 360° panoramic view. This procedure reproduces the characteristics of distortions observed in consumer-grade 360° captures. Consequently, the synthesized outputs accurately reflect these artifacts, making them suitable as imperfect ground truth for training and evaluating the 3D Gaussian splatting pipeline.

Moreover, once training is complete, our model can also render seamless 360° panoramas directly without the additional computational cost of the distortion process. Since the initial Gaussians are defined in an ideally calibrated space, omitting the distortion step during inference produces final renderings that closely approximate an ideal, distortion-free 360° image. In effect, the output resembles a 360° capture obtained even when the 3D Gaussian splatting is trained on imperfectly stitched inputs, effectively overcoming the spatial inaccuracies.

\section{Experiments}

\subsection{Experimental Setup}

\noindent{\bf Datasets.}
We evaluated camera calibration performance using a synthetic dataset derived from OmniBlender~\cite{choi2023egonerf}, designed to emulate typical distortions in dual-fisheye 360° camera setups. We generated equirectangular images by combining two virtual fisheye views (front and back), each rendered with a slightly displaced camera. The inter-camera displacement was randomly sampled between -2 and 2 cm in all directions. Each pair of fisheye images covered a 185° field of view and was stitched into a single distorted panorama to simulate realistic lens imperfections. Each synthetic scene included 50 equirectangular images, evenly split into 25 for training and 25 for testing.

To validate whether our method achieves seamless omnidirectional synthesis across various real-world scenarios, we conducted experiments on two real-world 360° datasets, following the evaluation protocol from OmniGS~\cite{li2024omnigs}. Specifically, we used the Ricoh360 dataset introduced in EgoNeRF~\cite{choi2023egonerf}, consisting of scenes captured with a Ricoh360 camera, and the 360Roam dataset~\cite{huang2022360roam}, captured using an Insta360 camera. Images from the 360Roam dataset were resized to $1520\times760$, and the lower region (below pixel row 712) was cropped to remove a mobile robot from the scene. To ensure broad applicability, our method operates directly on ERP-format panoramas, which are the standard for most publicly available 360° content.

\vspace{0.2cm}\noindent{\bf Metrics.}
We evaluated omnidirectional novel view synthesis using Peak Signal-to-Noise Ratio (PSNR), Structural Similarity Index (SSIM), and LPIPS (using AlexNet)~\cite{zhang2018perceptual, alex2017alexnet}. Additionally, we quantified camera positional calibration accuracy by computing the Mean Absolute Error (MAE) between the predicted and actual inter-camera gaps. 

\vspace{0.2cm}\noindent{\bf Implementations.}
Implementation of Seamless 360° Gaussian Splatting was performed using the PyTorch framework \cite{paszke2019pytorch} and the 360° rasterizer from OmniGS \cite{li2024omnigs}, which is built upon the original 3D Gaussian splatting framework \cite{kerbl20233dgs}. Experiments on real-world scenes adhered to the default parameter settings of the original 3D Gaussian splatting implementation, with one modification: the camera extent for the Ricoh360 dataset from EgoNeRF \cite{choi2023egonerf} was scaled by a factor of 10 to accommodate the limited camera motion. To ensure a fair comparison, all baseline models were trained using their officially released code on a single NVIDIA A6000 GPU with a training time limit of 90 minutes. Due to memory limitations on the Ricoh360 dataset, OP43DGS~\cite{li2024omnigs} utilized the checkpoint saved immediately before training terminated as a result of an out-of-memory error.
To train the learnable angular distortion grids \(\Delta\)$R_F,R_B$ and camera gaps \(\Delta\)$T_F,T_B$, we initialized the learning rate at 0.001 and gradually decayed it exponentially to 0.0001. The learning rate scheduling followed the approach used in JaxNeRF~\cite{jaxnerf2020github}, ensuring stable convergence while refining camera parameters throughout training. Additionally, a total variation loss was applied to 
\(\Delta\)$R_F,R_B$ to enforce smoothness and prevent abrupt variations in the estimated rotations.


\vspace{0.2cm}
\subsection{Synthetic Calibration Accuracy Assessment} 
We present a quantitative analysis of our model's calibration performance using synthetic scenes. Our dataset provides ground-truth camera gap values between the dual fisheye cameras, enabling a precise evaluation of positional accuracy. 
For comparison, we benchmark against OmniNeRF—a self-calibrating method capable of correcting both lens distortion and inter-camera gaps. Despite training OmniNeRF for 90 minutes—more than twice the duration required by our approach—as shown in Table~\ref{tab:GS_T_calibration_}, our method achieves an average positional error of only 0.11 cm. Our algorithm not only demonstrates exceptional calibration accuracy but also ensures precise scene reconstruction, as validated by the high-fidelity renderings in \cref{Fig:GS_comparison_quailitative} (top portion).
\begin{table}[ht!]
\vspace{-0.4cm}
\centering
\label{tab:metrics}
\resizebox{0.45\textwidth}{!}{ 
\begin{tabular}{lccccc}
\toprule
\multirow{2}{*}{\centering Method} & \multicolumn{5}{c}{Mean Absolute Error (cm)} \\
\cmidrule(lr){2-6}
& \textit{barbershop} & \textit{classroom} & \textit{restroom} & \textit{lone\_monk} & Average \\
\midrule
Ours                            & 0.1166 & 0.1965 & 0.1044 & 0.0423 & 0.1150 \\
OmniNeRF~\cite{gu2022omninerf}   & 2.761  & 3.602  & 3.454  & 3.377  & 3.299  \\
\bottomrule
\end{tabular}
}
\vspace{-0.2cm}
\caption{Inter-camera gap calibration performance comparison with OmniNeRF on the synthetic dataset. OmniNeRF utilizes dual fisheye images for self-calibration, and the calibration accuracy is reported as mean absolute error (MAE) in centimeters.}
\label{tab:GS_T_calibration_}
\vspace{-0.2cm}
\end{table}

\subsection{Quantitative and Qualitative Evaluation on Synthetic and Real-World Datasets}
We conducted extensive experiments to compare our method with state-of-the-art algorithms for 360° novel view synthesis \cite{choi2023egonerf, wang2024op43dgs, lee2024odgs, li2024omnigs} on both synthetic and real-world data. For the real-world evaluation, we analyzed 11 scenes from the EgoNeRF dataset (captured with a Ricoh360 camera) and 10 scenes from the 360Roam dataset (captured with an Insta360 camera). 
\begin{table}[b]
\vspace{-0.4cm}
\centering
\huge
\resizebox{1.0\linewidth}{!}{ 
    \begin{tabular}{llccccc}
    \toprule
    \textbf{Dataset}      & \textbf{Metric}    & \textbf{EgoNeRF} \cite{choi2023egonerf} & \textbf{OP43DGS} \cite{wang2024op43dgs} & \textbf{ODGS} \cite{lee2024odgs} & \textbf{OmniGS} \cite{li2024omnigs} & \textbf{Ours}   \\
    \midrule
\multirow{4}{*}{Synthetic} 
    & PSNR↑      & 25.468  & 27.825  & 28.833 & 32.421 & \textbf{34.986} \\
    & SSIM↑      & 0.746   & 0.874   & 0.866  & 0.927  & \textbf{0.947} \\
    & LPIPS↓     & 0.232   & 0.116   & 0.116  & 0.053  & \textbf{0.048} \\
    & \#Points   & -       & 411K    & 1002K   & 374K   & 308K    \\
\midrule
\multirow{4}{*}{Ricoh360} 
    & PSNR↑      & 24.639  & 23.778  & 22.556 & 26.064 & \textbf{26.960} \\
    & SSIM↑      & 0.746   & 0.749   & 0.739  & 0.829  & \textbf{0.852} \\
    & LPIPS↓     & 0.250   & 0.267   & 0.238  & 0.125  & \textbf{0.105} \\
    & \#Points   & -       & 329K    & 1276K  & 619K   & 621K    \\
\midrule
\multirow{4}{*}{360Roam} 
    & PSNR↑      & 23.021  & 23.689  & 22.394 & 25.083 & \textbf{26.101} \\
    & SSIM↑      & 0.693   & 0.760   & 0.710  & 0.803  & \textbf{0.829} \\
    & LPIPS↓     & 0.368   & 0.234   & 0.264  & 0.145  & \textbf{0.144} \\
    & \#Points   & -       & 236K    & 971K   & 836K   & 447K    \\
    \bottomrule
    \end{tabular}
}
\vspace{-0.2cm}
\caption{Quantitative comparisons on synthetic and real-world datasets. The table summarizes the average PSNR, SSIM, LPIPS, and point count across all evaluated datasets, highlighting the performance of our novel view synthesis method.}
\label{tab:GC_main_comparison}
\vspace{-0.4cm}
\end{table}

\begin{figure*}
    \centering
    \includegraphics[width=0.91\linewidth, trim=0 59 15 0, clip]{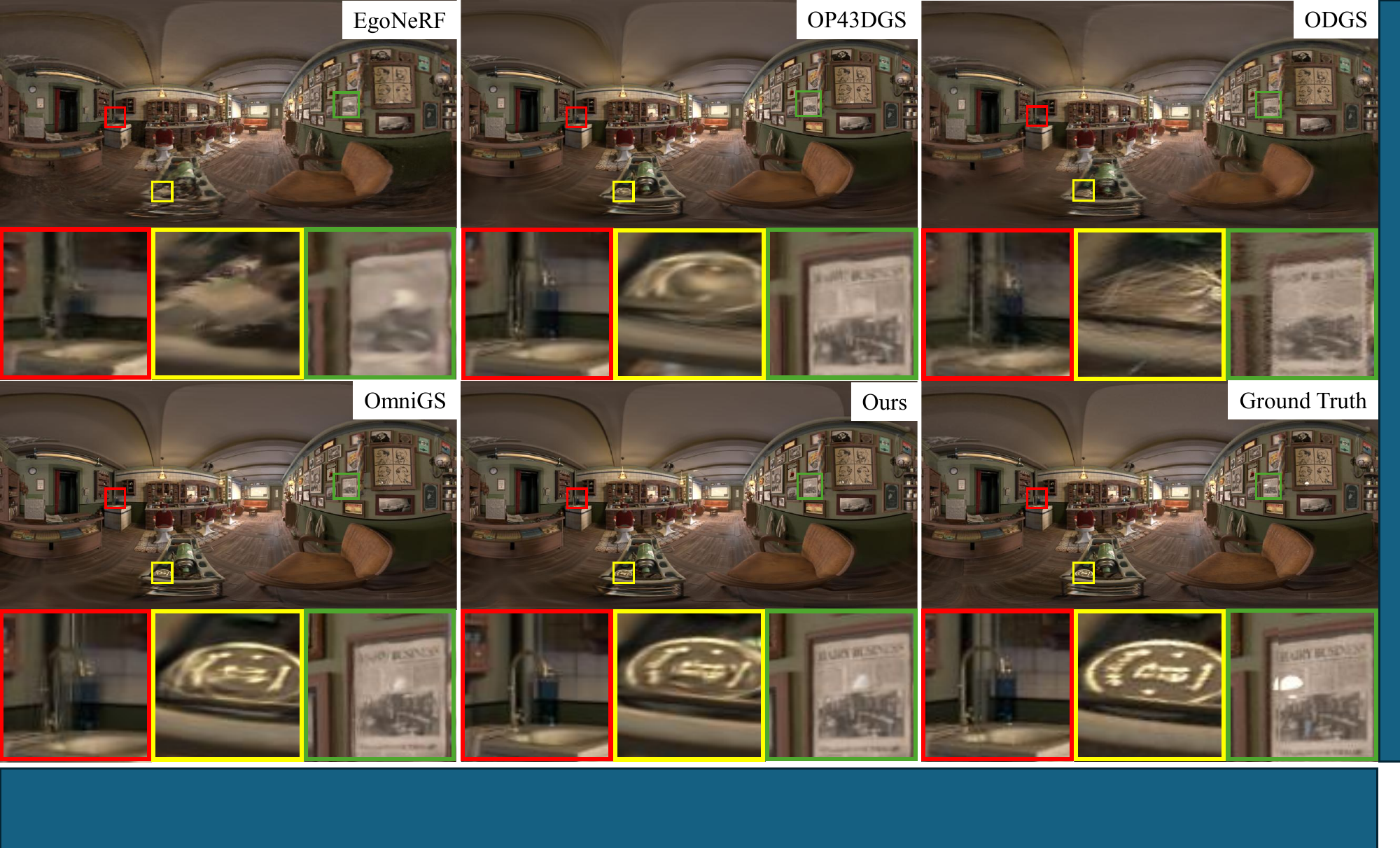}
    \vskip 1mm
    \includegraphics[width=0.91\linewidth, trim=0 51 15 0, clip]{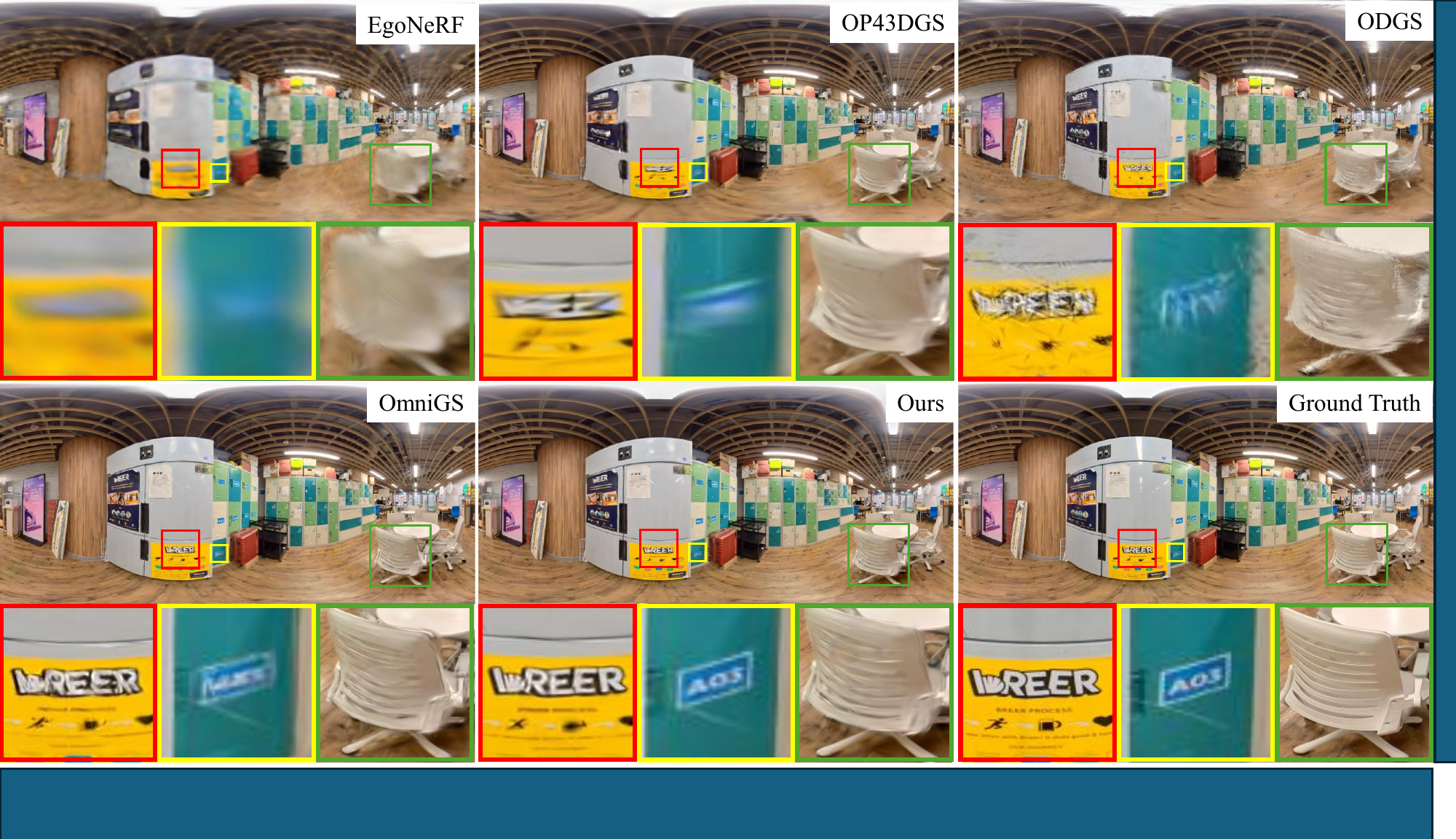}
    \caption{Qualitative comparisons of novel-view synthesis results. The top portion shows images from the synthetic dataset (\textit{barbershop} scene), while the bottom portion displays results from the 360Roam dataset (\textit{base} scene).}
    \label{Fig:GS_comparison_quailitative}
    \vspace{-0.4cm}
\end{figure*}
As summarized in Table~\ref{tab:GC_main_comparison}, our model outperforms competing methods across all metrics while utilizing fewer Gaussians, with particularly notable gains on indoor scenes from the 360Roam dataset.( Full experimental results are provided in the supplementary material )

\begin{table*}[t]
\centering
\resizebox{0.95\linewidth}{!}{%
\begin{tabular}{lcccc}
\toprule
\textbf{Scene} & \textbf{Ours} & \textbf{Ours w/o Lens Calib} & \textbf{Ours w/o Gap} & \textbf{Ours w/o Lens Calib, Gap}\\
\midrule
\textit{ } & PSNR↑ / SSIM↑ / LPIPS↓ / \#Points  & PSNR↑ / SSIM↑ / LPIPS↓ / \#Points  & PSNR↑ / SSIM↑ / LPIPS↓ / \#Points & PSNR↑ / SSIM↑ / LPIPS↓ / \#Points \\
\midrule
\textit{Bar} & \textbf{23.155} / \textbf{0.804} / \textbf{0.157} / 592K  & 22.593 / 0.785 / 0.167 / 587K  & 23.040 / 0.801 / 0.158 / 585K & 22.497 / 0.782 / 0.168 / 587K \\
\textit{Base} & \textbf{26.496} / \textbf{0.856} / \textbf{0.089} / 806K  & 25.487 / 0.823 / 0.101 / 804K  & 26.295 / 0.852 / 0.090 / 796K & 24.918 / 0.805 / 0.108 / 788K \\
\textit{Cafe} & \textbf{26.139} / \textbf{0.859} / \textbf{0.099} / 518K  & 25.472 / 0.841 / 0.108 / 556K  & 26.048 / 0.857 / 0.100 / 513K & 25.231 / 0.835 / 0.112 / 551K \\
\textit{Canteen} & \textbf{22.863} / \textbf{0.762} / \textbf{0.204} / 322K  & 22.652 / 0.754 / 0.208 / 326K  & 22.813 / 0.760 / \textbf{0.204} / 317K & 22.443 / 0.744 / 0.211 / 314K \\
\textit{Center} & \textbf{25.824} / \textbf{0.822} / \textbf{0.176} / 372K  & 25.312 / 0.808 / 0.181 / 400K  & 25.747 / 0.820 / 0.177 / 371K & 25.135 / 0.804 / 0.184 / 395K \\
\textit{Corridor} & \textbf{25.733} / \textbf{0.826} / \textbf{0.166} / 202K  & 25.381 / 0.818 / 0.173 / 205K & 25.135 / 0.819 / 0.169 / 199K & 25.238 / 0.810 / 0.174 / 206K \\
\textit{Innovation} & \textbf{27.174} / \textbf{0.852} / \textbf{0.120} / 784K  & 26.364 / 0.821 / 0.129 / 811K  & 27.084 / 0.851 / \textbf{0.120} / 784K & 26.053 / 0.811 / 0.133 / 801K \\
\textit{Lab} & \textbf{30.171} / \textbf{0.917} / \textbf{0.060} / 329K  & 28.888 / 0.900 / 0.069 / 351K  & 29.946 / 0.914 / 0.061 / 327K & 28.501 / 0.895 / 0.072 / 346K \\
\textit{Library} & \textbf{27.236} / \textbf{0.785} / \textbf{0.191} / 283K  & 26.641 / 0.767 / 0.197 / 293K  & 27.187 / 0.784 / 0.192 / 281K & 26.326 / 0.760 / 0.200 / 291K \\
\textit{Office} & \textbf{26.212} / \textbf{0.804} / \textbf{0.180} / 265K  & 25.746 / 0.795 / 0.184 / 277K  & 25.912 / 0.801 / 0.183 / 262K & 25.022 / 0.783 / 0.195 / 269K \\
\textbf{\textit{Average}}      & \textbf{26.101} / \textbf{0.829} / \textbf{0.144} / 447K  & 25.453 / 0.811 / 0.152 / 461K  & 25.921 / 0.826 / 0.145 / 444K & 25.136 / 0.803 / 0.156 / 455K \\

\bottomrule
\end{tabular}
}
\caption{Ablation study evaluating the effectiveness of two key components in our training pipeline: translation-based adjustments for simulating camera gap and angular distortion-based transformations for modeling lens distortion. "w/o gap" indicates the removal of the camera gap component, and "w/o lens calib" denotes the exclusion of the lens distortion component. Removing these elements results in a significant performance degradation, particularly underscoring the strong impact of angular distortion correction.}

\label{tab:GS_ablation2_roam}
\vspace{-0.3cm}
\end{table*}

Qualitative evaluation further supports these findings. As shown in \cref{Fig:GS_comparison_quailitative}, our model produces the highest quality outputs, while other state-of-the-art 360° Gaussian Splatting algorithms struggle to capture fine details and accurately reproduce text. As depicted in \cref{Fig:figure0_headline}, camera gap and lens errors lead to stitching issues—resulting in artifacts such as ringing effects, object overlap, and blurriness due to imprecise ground truth. 

Additionally, \cref{Fig:GS_figure2_seamless} presents a comparison between seamlessly rendered images and realistic results from real-world scenes. The ground-truth image exhibits noticeable stitching errors along the seam due to the dual fisheye configuration, whereas our seamless model reconstructs the scene with higher fidelity—sometimes even outperforming the captured images.

We also compare training time, peak VRAM usage, and inference speed against OmniGS (see Table~\ref{tab:tab4_traintime}). Although our approach requires more than twice the training time and up to 2 GB additional VRAM, it achieves 30\% faster rendering at test time. This speedup stems from our stitching model’s precise Gaussian fitting, which generates accurate camera rays, reducing the number of primitives needed for high-quality novel-view synthesis.
\begin{table}[!t]
\centering
\resizebox{0.92\linewidth}{!}{
    \begin{tabular}{clcccc}
    \toprule
    \textbf{Method} & \textbf{Scene} & \textbf{Train Time} & \textbf{Peak Mem} & \textbf{FPS (Test)} & \textbf{\#Points} \\
    \midrule
    \multirow{2}{*}{OmniGS} & Office  & 20\,m\,31\,s & 5\,679\,MB & 96.52 & 600K \\
                            & Library & 21\,m\,22\,s & 5\,140\,MB & 83.17 & 526K \\
    \addlinespace
    \multirow{2}{*}{Ours}   & Office  & 54\,m\,47\,s & 6\,633\,MB & 136.12 & 265K \\ 
                            & Library & 56\,m\,49\,s & 7\,143\,MB & 112.68 & 283K \\
    \bottomrule
    \end{tabular}
}
\vspace{-0.2cm}
\caption{Comparison of training time, VRAM usage, and test-time FPS between OmniGS~\cite{li2024omnigs} and our method on 360Roam scenes (Office, Library).}
\vspace{-0.2cm}
\label{tab:tab4_traintime}
\end{table}
\subsection{Ablation Study on Our Seamless Rendering Components}

\begin{figure}
    \centering
    \includegraphics[width=0.86\linewidth, trim=0 70 580 10, clip]{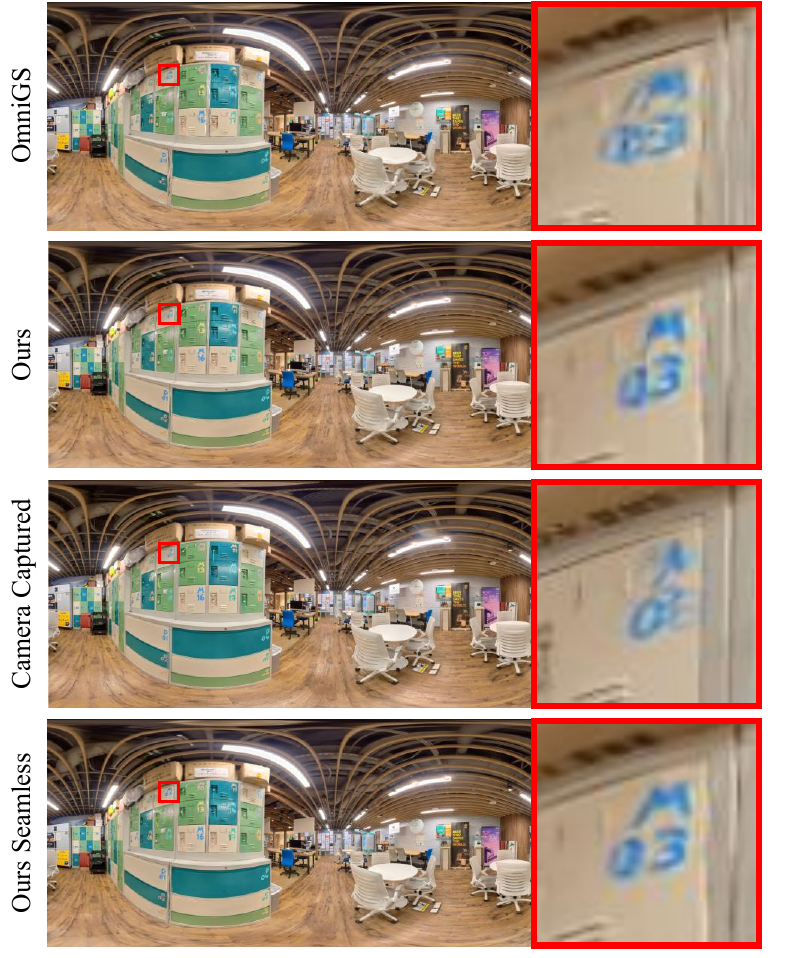}
    \vspace{-3pt}
    \caption{Qualitative comparison on the 360Roam \textit{base}. The top row displays the output of the state-of-the-art model OmniGS, followed by our method's results, the ground truth, and finally the seamless rendering output produced by our method.}
    \label{Fig:GS_figure2_seamless}
    \vspace{-0.4cm}
\end{figure}

To evaluate our calibration process, we conducted ablation experiments using the 360Roam indoor dataset (see \cref{Fig:GS_figure2_seamless}), where calibration is particularly critical due to the prevalence of close-range objects. Our model adjusts Gaussian splats via two mechanisms—translation and angular distortion—thereby simulating both camera gap and lens distortion during training. We assessed the contribution of each component with experiments summarized in Tab.~\ref{tab:GS_ablation2_roam}. The results indicate that both mechanisms significantly enhance performance, with angular distortion having an especially strong effect; notably, disabling both adjustments results in an average PSNR drop of 1 dB.

\section{Conclusion}
In this work, we introduced an integrated calibration framework for omnidirectional novel view synthesis that addresses the challenges inherent in dual-fisheye camera systems. By incorporating translation-based adjustments and angular distortion corrections within the 360° Gaussian Splatting paradigm, our method effectively compensates for camera gap and lens distortions, leading to highly accurate scene reconstructions. Extensive experiments on both synthetic and real-world datasets demonstrate that our approach not only outperforms state-of-the-art methods in terms of quantitative metrics such as PSNR, SSIM, and LPIPS, but also produces superior qualitative results with enhanced detail and fewer artifacts. Our seamless rendering pipeline, validated through rigorous ablation studies, underscores the importance of precise calibration in achieving high-fidelity outputs. Future work will focus on further refining our calibration techniques, exploring real-time applications, and extending the framework to more complex imaging scenarios.

\vspace{0.2cm} \noindent{\bf Limitations} 
A notable limitation of our approach is reduced performance in textureless regions. Uniform walls and smooth areas (e.g., the sky) lack sufficient features for reliable calibration, and simultaneous view synthesis and calibration further exacerbate errors.

{
    \clearpage
    \small
    \section*{Acknowledgements} This work was supported by the Institute of Information \& Communications Technology Planning \& Evaluation (IITP) with a grant funded by the Ministry of Science and ICT (MSIT) of the Republic of Korea in connection with the Global AI Frontier Lab International Collaborative Research. (No. RS-2024-00469482 \& RS-2024-00509279), Artificial Intelligence Graduate School Program, Yonsei University, under Grant RS-2020-11201361.
    
    \bibliographystyle{ieeenat_fullname}
    \bibliography{main}
}

\maketitlesupplementary
\setcounter{section}{0}
\renewcommand\thesection{\Alph{section}}
\section{Baseline Comparisons with Dual-Fisheye and MVG}
\begin{figure}[htbp]
  \centering
  \begin{minipage}{0.7\linewidth}
    \centering
    \captionof{table}{Quantitative comparison on raw fisheye inputs.}
    \vspace{-0.7em}
    \label{tab:fisheye}
    \resizebox{0.9\linewidth}{!}{
      \begin{tabular}{cccc ccc}
        \toprule
        & \multicolumn{3}{c}{Vanilla} & \multicolumn{3}{c}{Ours} \\
        \cmidrule(lr){2-4} \cmidrule(lr){5-7}
        Scene & PSNR↑ & LPIPS↓ & \#Points & PSNR↑ & LPIPS↓ & \#Points \\
        \midrule
        Barbershop & 29.9564 & 0.0553 & 507k & 30.2923 & 0.0530 & 479k \\
        \bottomrule
      \end{tabular}
    }
  \end{minipage}%
  \vspace{0.35em} 
  \begin{minipage}{0.6\linewidth}
    \centering
    \includegraphics[width=0.97\linewidth,trim=0 160 80 0,clip]{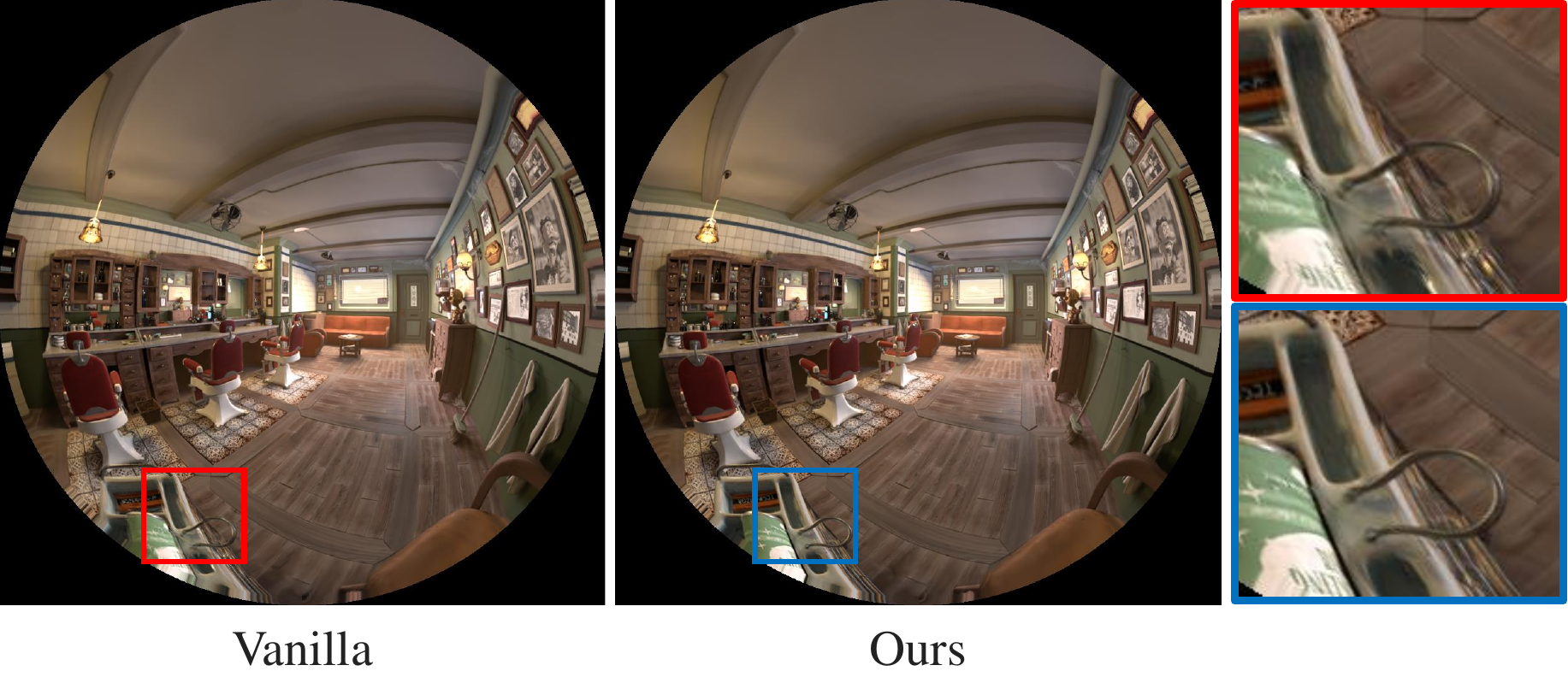}
     \vspace{-0.75em}
    \caption{Qualitative comparison on raw fisheye inputs: vanilla fisheye model (Vanilla) vs.\ our seamless fisheye model (Ours).}
    \vspace{-0.4cm}
    \label{fig:fisheye}
  \end{minipage}
\end{figure}

We further evaluate our method using raw dual-fisheye inputs. Tab.\ref{tab:fisheye} and Fig.\ref{fig:fisheye} show that applying our seamless calibration algorithm to fisheye images yields a meaningful improvement (PSNR +0.34dB, LPIPS -0.0023), comparable to results obtained with ERP inputs. This demonstrates that our pipeline is readily applicable to alternative image formats.
We also implemented a distortion-aware MVG pipeline for comparison. However, we observed significantly worse 360° rendering performance than with the ERP-based rasterizer. This aligns with previous findings from works such as OmniGS and OP43DGS, which consistently show that ERP-based rasterization outperforms MVG approaches in novel-view synthesis, even without explicit distortion correction. By adopting ERP inputs, our method builds on a strong baseline while avoiding the complexity of distortion-correction preprocessing.

\section{Modeling of Angular Distortion Grid}

To evaluate the impact of the embedding grid resolution, we conducted experiments with resolutions of $16 \times 32$, $32 \times 64$, $64 \times 128$, and $128 \times 256$. The results showed nearly identical performance across all resolutions, with differences in PSNR remaining below $0.2\,\mathrm{dB}$, which demonstrates the robustness of our method to the embedding resolution. We attribute this stability to the smooth variation of angular distortion across the sphere, allowing the embedding to generalize well even at relatively low resolutions. Interestingly, we observe that \(\Delta R\) learned from different scenes of the same camera are nearly identical, supporting cross-scene transferability. However, due to significant inter-device calibration differences, cross-device transfer remains challenging.

\section{Additional Quantitative and Qualitative Evaluation}
\label{sec:suppl_sec1}
In the supplementary material, we provide comprehensive experimental results covering quantitative evaluations, qualitative assessments, and seamless rendering performance. Quantitative results detail our method's superior performance in metrics such as PSNR, SSIM, and LPIPS, while also highlighting its efficiency with fewer Gaussians. Qualitative evaluations showcase the method's ability to capture fine details and maintain visual fidelity across challenging scenes. Additionally, seamless rendering results demonstrate the effectiveness of our calibration in eliminating stitching artifacts and producing artifact-free, high-fidelity outputs even when the ground truth suffers from noticeable errors.

\begin{table*}[h]
    \centering
    \small
    \renewcommand{\arraystretch}{1.0}
    \resizebox{0.84\textwidth}{!}{
        \begin{tabular}{llccccc}
            \toprule
            \textbf{Scene} & \textbf{Method} &  \textbf{EgoNeRF \cite{choi2023egonerf}} & \textbf{OP43DGS \cite{wang2024op43dgs}} & \textbf{ODGS \cite{lee2024odgs}} & \textbf{OmniGS \cite{li2024omnigs}} & \textbf{Ours} \\
            \midrule
\multirow{4}{*}{\textit{bricks}} & PSNR↑  & 22.708 & 21.270  & 21.608 & 24.514 & \textbf{26.182}\\
            & SSIM↑  & 0.722  & 0.705  & 0.744  & 0.840  & \textbf{0.882}  \\
            & LPIPS↓ & 0.246  & 0.268  & 0.190  & 0.112  & \textbf{0.079}  \\
            & \#Points & - & 202K & 2075K & 914K & 900K \\
\midrule
\multirow{4}{*}{\textit{bridge}} & PSNR↑  & 22.973 & 22.528 & 21.901 & 23.650 & \textbf{24.316}\\
            & SSIM↑  & 0.715  & 0.740  & 0.712  & 0.795  & \textbf{0.818}  \\
            & LPIPS↓ & 0.258  & 0.217  & 0.243  & 0.133  & \textbf{0.107}  \\
            & \#Points & - & 721K & 1311K & 758K & 840K \\
\midrule
\multirow{4}{*}{\textit{bridge\_under}} & PSNR↑  & 24.213 & 22.754  & 23.599 & 26.696 & \textbf{27.096}\\
            & SSIM↑  & 0.764  & 0.738  & 0.777  & 0.877  & \textbf{0.895}  \\
            & LPIPS↓ & 0.232  & 0.292  & 0.200  & 0.086  & \textbf{0.072}  \\
            & \#Points & - & 222K & 1876K & 815K & 878K \\
\midrule
\multirow{4}{*}{\textit{cat\_tower}} & PSNR↑  & 23.715 & 23.592 & 22.735 & 24.775 & \textbf{25.571}\\
            & SSIM↑  & 0.684  & 0.717  & 0.696  & 0.775  & \textbf{0.795}  \\
            & LPIPS↓ & 0.299  & 0.269  & 0.242  & 0.155  & \textbf{0.144}  \\
            & \#Points & - & 486K & 1177K & 548K & 529K \\
\midrule
\multirow{4}{*}{\textit{center}} & PSNR↑  & 28.065 & 27.616 & 23.167 & 29.381 & \textbf{30.749}\\
            & SSIM↑  & 0.850  & 0.858  & 0.776  & 0.893  & \textbf{0.912}  \\
            & LPIPS↓ & 0.189  & 0.181  & 0.321  & 0.094  & \textbf{0.074}  \\
            & \#Points & - & 293K & 234K & 353K & 356K \\
\midrule
\multirow{4}{*}{\textit{farm}} & PSNR↑  & 21.938 & 21.214 & 20.865 & 22.210 & \textbf{23.067}\\
            & SSIM↑  & 0.651  & 0.663  & 0.659  & 0.732  & \textbf{0.773}  \\
            & LPIPS↓ & 0.299  & 0.286  & 0.238  & 0.157  & \textbf{0.130}  \\
            & \#Points & - & 478K & 2110K & 957K & 900K \\
\midrule
\multirow{4}{*}{\textit{flower}} & PSNR↑  & 21.520 & 19.634 & 18.752 & 22.389 & \textbf{23.089}\\
            & SSIM↑  & 0.620  & 0.568  & 0.591  & 0.728  & \textbf{0.754}  \\
            & LPIPS↓ & 0.341  & 0.548  & 0.334  & 0.187  & \textbf{0.166}  \\
            & \#Points & - & 55K & 1458K & 664K & 637K \\
\midrule
\multirow{4}{*}{\textit{gallery\_chair}} & PSNR↑  & 27.053 & 27.320 & 25.761 & 28.685 & \textbf{28.971}\\
            & SSIM↑  & 0.834  & 0.858  & 0.828  & 0.892  & \textbf{0.905}  \\
            & LPIPS↓ & 0.227  & 0.199  & 0.217  & 0.107  & \textbf{0.088}  \\
            & \#Points & - & 325K & 628K & 329K & 330K \\
\midrule
\multirow{4}{*}{\textit{gallery\_pillar}} & PSNR↑  & 26.764 & 27.121 & 22.493 & 28.712 & \textbf{29.855}\\
            & SSIM↑  & 0.822  & 0.850  & 0.805  & 0.885  & \textbf{0.904}  \\
            & LPIPS↓ & 0.180  & 0.155  & 0.191  & 0.092  & \textbf{0.073}  \\
            & \#Points & - & 401K & 764K & 374K & 419K \\
\midrule
\multirow{4}{*}{\textit{garden}} & PSNR↑  & 26.483 & 24.779 & 22.576 & 27.170 & \textbf{27.815}\\
            & SSIM↑  & 0.715  & 0.728  & 0.717  & 0.798  & \textbf{0.817}  \\
            & LPIPS↓ & 0.276  & 0.276  & 0.234  & 0.145  & \textbf{0.133}  \\
            & \#Points & - & 293K & 1042K & 499K & 537K \\
\midrule
\multirow{4}{*}{\textit{poster}} & PSNR↑  & 25.598 & 23.738 & 24.662 & 28.518 & \textbf{29.851}\\
            & SSIM↑  & 0.832  & 0.808  & 0.825  & 0.902  & \textbf{0.915}  \\
            & LPIPS↓ & 0.200  & 0.241  & 0.212  & 0.105  & \textbf{0.092}  \\
            & \#Points & - & 138K & 1362K & 593K & 505K \\
            \bottomrule
        \end{tabular}
    }
    \caption{Per-scene quantitative evaluation results on EgoNeRF-Ricoh360 dataset.}
    \label{tab:GS_comparison_ricoh_}
\end{table*}

\begin{table*}[h]
    \centering
    \small
    \renewcommand{\arraystretch}{1.0}
    \resizebox{0.9\textwidth}{!}{ 
        \begin{tabular}{llcccccc}
            \toprule
            \textbf{Scene} & \textbf{Method} &  \textbf{EgoNeRF \cite{choi2023egonerf}} & \textbf{OP43DGS \cite{wang2024op43dgs}} & \textbf{ODGS \cite{lee2024odgs}} & \textbf{OmniGS \cite{li2024omnigs}} & \textbf{SC-OmniGS \cite{huang2025scomnigs}} & \textbf{Ours} \\
            \midrule
\multirow{4}{*}{\textit{Bar}}   & PSNR↑  & 19.689 & 20.947 & 19.812 & 22.536 & 22.556 & \textbf{23.155}\\
                                & SSIM↑  & 0.619  & 0.715  & 0.666  & 0.783  & 0.783 & \textbf{0.804}\\
                                & LPIPS↓ & 0.424  & 0.268  & 0.277  & 0.160  & 0.200 & \textbf{0.157}\\
                                & \#Points & - & 215K & 1244K & 1104K & - & 592K \\
\midrule
\multirow{4}{*}{\textit{Base}}  & PSNR↑  & 21.606 & 23.367 & 22.152 & 24.924 & 25.504 & \textbf{26.496}\\
                                & SSIM↑  & 0.608  & 0.745  & 0.679  & 0.806  & 0.816  & \textbf{0.856}\\
                                & LPIPS↓ & 0.434  & 0.206  & 0.230  & 0.099  & 0.133  & \textbf{0.089}\\
                                & \#Points & - & 326K & 1428K & 1236K & - & 806K \\
\midrule
\multirow{4}{*}{\textit{Cafe}}  & PSNR↑  & 22.323 & 23.432 & 22.457 & 25.307 & 24.838 & \textbf{26.139}\\
                                & SSIM↑  & 0.702  & 0.793  & 0.738  & 0.838  & 0.813  & \textbf{0.859}\\
                                & LPIPS↓ & 0.342  & 0.190  & 0.208  & 0.101  & 0.161  & \textbf{0.099}\\
                                & \#Points & - & 339K & 1389K & 1032K & - & 518K \\
\midrule
\multirow{4}{*}{\textit{Canteen}} & PSNR↑  & 20.960 & 21.313 & 20.194 & 22.458 & 22.159 & \textbf{22.863}\\
                                & SSIM↑  & 0.653  & 0.707  & 0.658  & 0.743  & 0.734  & \textbf{0.762}\\
                                & LPIPS↓ & 0.439  & 0.294  & 0.304  & \textbf{0.196} & 0.263 & 0.204\\
                                & \#Points & - & 164K & 854K & 686K & - & 322K \\
\midrule
\multirow{4}{*}{\textit{Center}}  & PSNR↑  & 23.041 & 24.144 & 22.768 & 25.121 & 25.802 & \textbf{25.824}\\
                                & SSIM↑  & 0.721  & 0.774  & 0.725  & 0.804  & 0.813 & \textbf{0.822}\\
                                & LPIPS↓ & 0.404  & 0.265  & 0.291  & \textbf{0.170}  & 0.203 & 0.176\\
                                & \#Points & - & 182K & 855K & 785K & - & 372K \\
\midrule
\multirow{4}{*}{\textit{Corridor}} & PSNR↑  & \textbf{26.644} & 24.097 & 23.606 & 25.037 & - & 25.733\\
                                & SSIM↑  & 0.788  & 0.790  & 0.740  & 0.812  & - & \textbf{0.826}\\
                                & LPIPS↓ & 0.255  & 0.243  & 0.287  & \textbf{0.162} & - & 0.166\\
                                & \#Points & - & 111K & 388K & 380K & - & 202K \\
\midrule
\multirow{4}{*}{\textit{Innovation}} & PSNR↑  & 23.643 & 24.358 & 22.492 & 25.902 & 26.390 & \textbf{27.174}\\
                                & SSIM↑  & 0.681  & 0.754  & 0.712  & 0.808  & 0.819  & \textbf{0.852}\\
                                & LPIPS↓ & 0.364  & 0.234  & 0.232  & \textbf{0.120}  & 0.148 & \textbf{0.120}\\
                                & \#Points & - & 296K & 1248K & 1305K & - & 784K \\
\midrule
\multirow{4}{*}{\textit{Lab}}    & PSNR↑  & 24.932 & 26.223 & 24.845 & 28.742 & 28.875 & \textbf{30.171}\\
                                & SSIM↑  & 0.791  & 0.859  & 0.820  & 0.897  & 0.898  & \textbf{0.917}\\
                                & LPIPS↓ & 0.263  & 0.127  & 0.168  & 0.063  & 0.087  & \textbf{0.060}\\
                                & \#Points & - & 372K & 1045K & 703K & - & 329K \\
\midrule
\multirow{4}{*}{\textit{Library}} & PSNR↑  & 23.381 & 25.393 & 23.818 & 25.946 & 26.250 & \textbf{27.236}\\
                                & SSIM↑  & 0.657  & 0.726  & 0.678  & 0.756  & 0.746 & \textbf{0.785}\\
                                & LPIPS↓ & 0.403  & 0.244  & 0.300  & 0.203  & 0.243 & \textbf{0.191}\\
                                & \#Points & - & 216K & 824K & 600K & - & 283K \\
\midrule
\multirow{4}{*}{\textit{Office}}  & PSNR↑  & 23.986 & 23.622 & 21.796 & 24.862 & - & \textbf{26.212}\\
                                & SSIM↑  & 0.714  & 0.739  & 0.681  & 0.787  & - & \textbf{0.804}\\
                                & LPIPS↓ & 0.348  & 0.269  & 0.344  & \textbf{0.175}  & - & 0.180\\
                                & \#Points & - & 138K & 434K & 526K & - & 265K \\
            \bottomrule
        \end{tabular}
    }
    \caption{Per-scene Quantitative evaluation results on the 360Roam dataset.}
    \label{tab:GS_comparison_roam_}
\end{table*}

\begin{table*}
    \centering
    \footnotesize
    \resizebox{0.9\linewidth}{!}{
        \begin{tabular}{llccccc}
            \toprule
            \textbf{Scene} & \textbf{Method} &  \textbf{EgoNeRF} \cite{choi2023egonerf} & \textbf{OP43DGS} \cite{wang2024op43dgs} & \textbf{ODGS} \cite{lee2024odgs} & \textbf{OmniGS} \cite{li2024omnigs} & \textbf{Ours} \\
            \midrule
            \multirow{4}{*}{\textit{barbershop}} & PSNR↑  & 28.150 & 29.703 & 28.844 & 33.283 & \textbf{36.022}\\
                        & SSIM↑  & 0.873 & 0.913 & 0.884 & 0.950  & \textbf{0.963}\\
                        & LPIPS↓ & 0.121 & 0.078 & 0.103 & 0.044  & \textbf{0.039}\\
                        & \#Points & - & 401K &	1002K & 316K & 242K \\
            \midrule
            \multirow{4}{*}{\textit{classroom}} & PSNR↑  & 28.788 &	31.016 & 30.655 & 34.547 & \textbf{36.362}\\
                        & SSIM↑  & 0.883 & 0.926 & 0.915 & 0.958 & \textbf{0.966}\\
                        & LPIPS↓ & 0.102 & 0.092 & 0.088 & 0.039  & \textbf{0.036}\\
                        & \#Points & - & 376K & 979K & 294K & 264K \\
            \midrule
            \multirow{4}{*}{\textit{restroom}} & PSNR↑  & 28.788 & 26.116 &	29.729 & 32.998 & \textbf{35.122}\\
                        & SSIM↑  & 0.772 & 0.820 & 0.818 & 0.896  & \textbf{0.922}\\
                        & LPIPS↓ & 0.256 & 0.177 & 0.169 & 0.082  & \textbf{0.076}\\
                        & \#Points & - & 378K & 840K & 402K & 333K \\
            \midrule
            \multirow{4}{*}{\textit{lone\_monk}} & PSNR↑  & 16.160 & 24.466 & 26.105 & 28.858 & \textbf{32.439}\\
            & SSIM↑  & 0.455	&0.837	&0.846	&0.906	&\textbf{0.939}  \\
            & LPIPS↓ & 0.450	&0.118	&0.104	&0.049	&\textbf{0.039}  \\
            & \#Points & - & 487K	&1189K	&483K	&394K \\
            \bottomrule
        \end{tabular}
        }
    \caption{Per-scene Quantitative evaluation results on the synthetic dataset.}
    \label{tab:GS_comparison_syn_}
\end{table*}
\clearpage

\begin{figure*}
    \centering
    \includegraphics[width=1\linewidth, trim=0 59 16 0, clip]{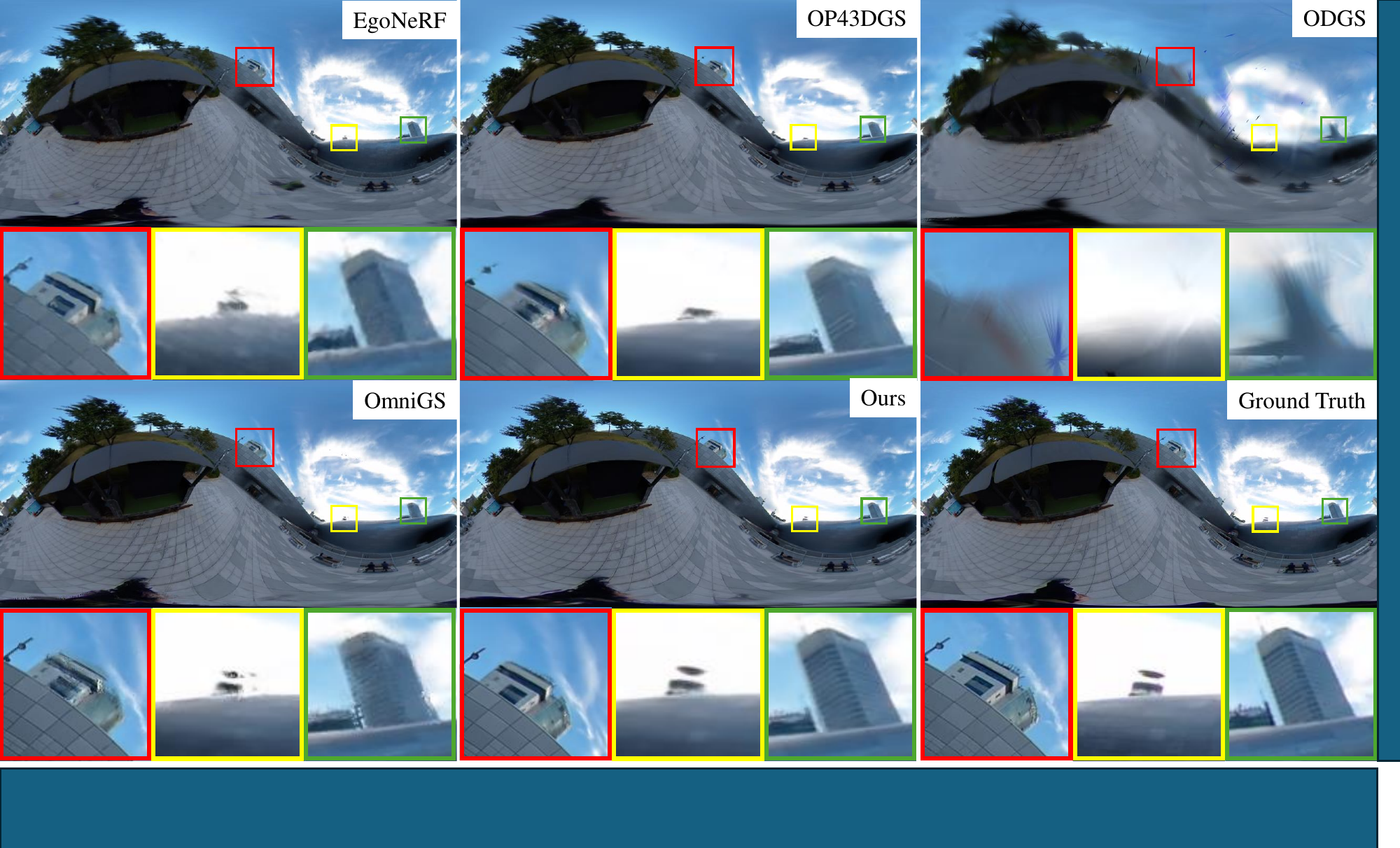}
    \caption{Qualitative comparisons of novel-view synthesis on \textit{center} scene in the EgoNeRF-Ricoh360 dataset.}
    \label{Fig:GS_comparison_ricoh_center}
\end{figure*}
\begin{figure*}
    \centering
    \includegraphics[width=1\linewidth, trim=0 59 16 0, clip]{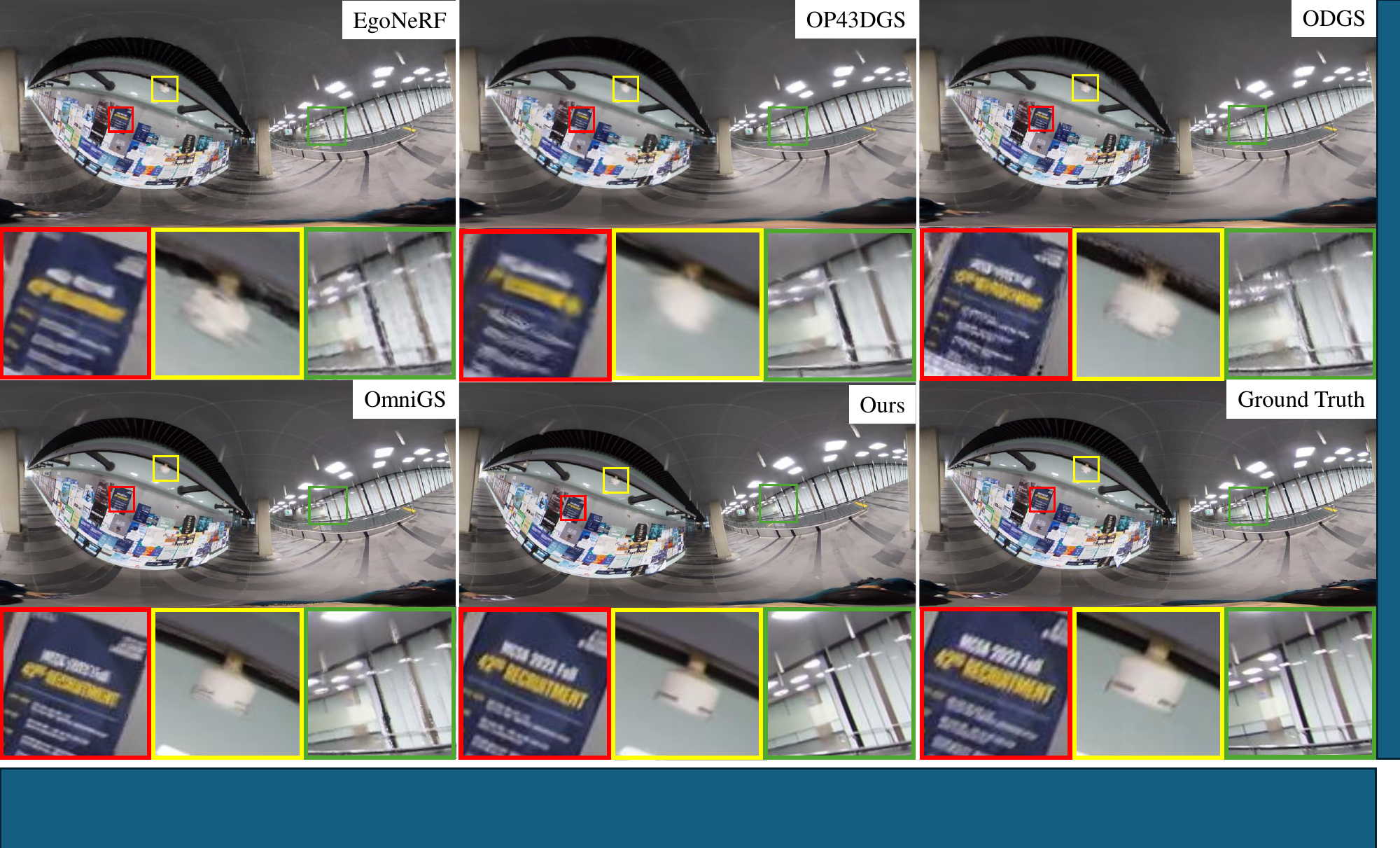}
    \caption{Qualitative comparisons of novel-view synthesis on \textit{poster} scene in the EgoNeRF-Ricoh360 dataset.}
    \label{Fig:GS_comparison_ricoh_poster}
\end{figure*}
\begin{figure*}
    \centering
    \includegraphics[width=1\linewidth, trim=0 51 15 0, clip]{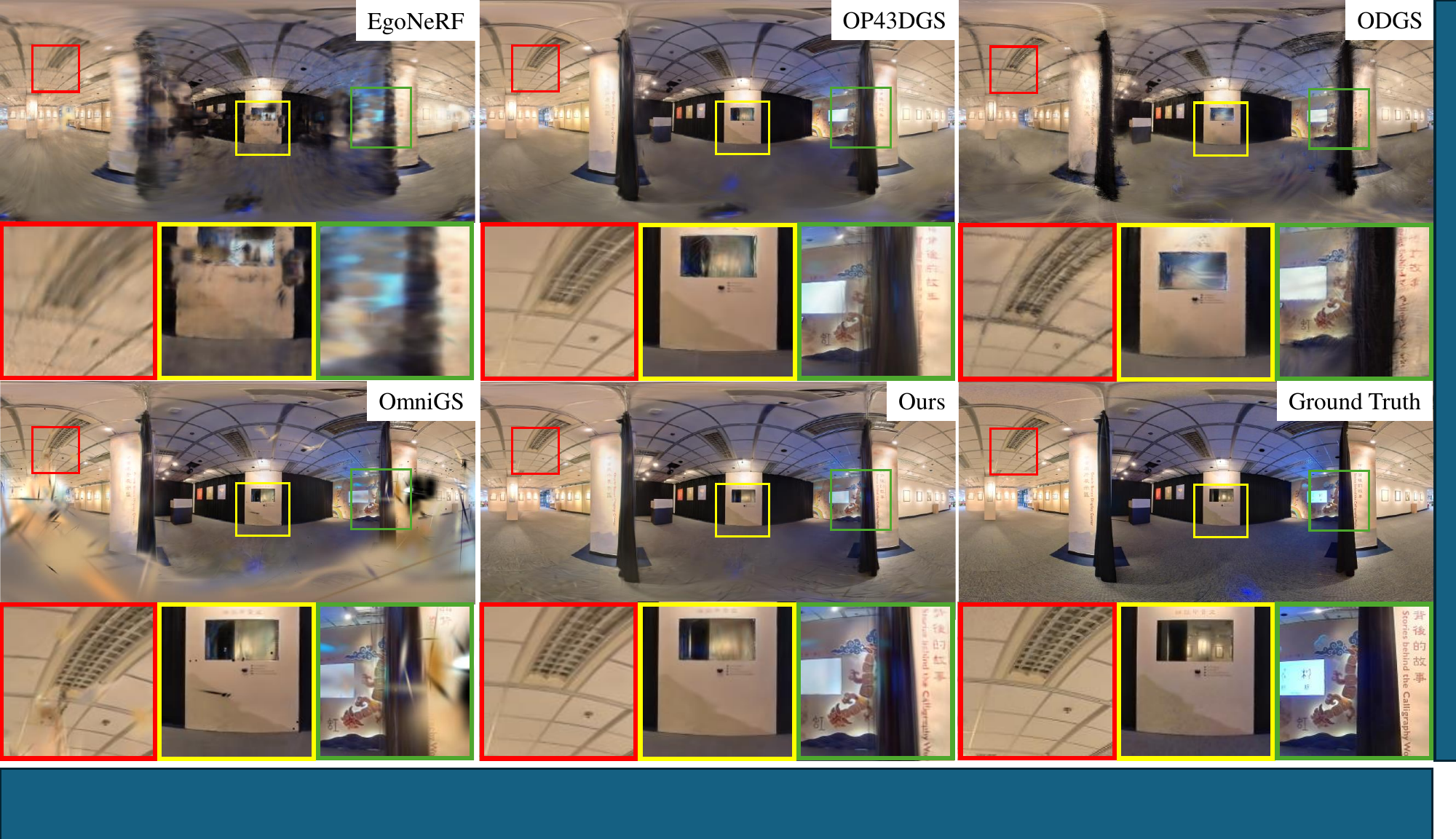}
    \caption{Qualitative comparisons of novel-view synthesis on \textit{library} scene in the 360Roam dataset.}
    \label{Fig:GS_comparison_roam_library}
\end{figure*}



\end{document}